# Automated Bi-Fold Weighted Ensemble Algorithms and its Application to Brain Tumor Detection and Classification

**PoTsang B. Huang**[2,5], **Muhammad Rizwan**[1,2,4,*], **and Mehboob Ali**[3,6]

[1] Department of Electrical Engineering and Computer Science, National Taiwan University, Taipei, Taiwan
[2] Department of Industrial and Systems Engineering, Chung Yuan Christian University, Zhongli, Taiwan.
[3] Department of Applied Mathematics, Chung Yuan Christian University, Zhongli, Taiwan.
[4]shahani753@gmail.com, [5] pthuang@cycu.edu.tw, [6]inform.me.8@gmail.com

## Abstract

The uncontrolled and unstructured growth of brain cells is known as brain tumor, which has one of the highest mortality rates among diseases from all types of cancers. Due to limited diagnostic and treatment capabilities, they pose significant challenges, especially in third-world countries. Early diagnosis plays a vital role in effectively managing brain tumors and reducing mortality rates. However, the availability of diagnostic methods is hindered by various limitations, including high costs and lengthy result acquisition times, impeding early detection of the disease. In this study, we present two cutting-edge bi-fold weighted voting ensemble models that aim to boost the effectiveness of weighted ensemble methods. These two proposed methods combine the classification outcomes from multiple classifiers and determine the optimal result by selecting the one with the highest probability in the first approach, and the highest weighted prediction in the second technique. These approaches significantly improve the overall performance of weighted ensemble techniques. In the first proposed method, we improve the soft voting technique (SVT) by introducing a novel unsupervised weight calculating schema (UWCS) to enhance its weight assigning capability, known as the extended soft voting technique (ESVT). Secondly, we propose a novel weighted method (NWM) by using the proposed UWCS. Both of our approaches incorporate three distinct models: a custom-built CNN, VGG-16, and InceptionResNetV2 which has been trained on publicly available datasets. The effectiveness of our proposed systems is evaluated through blind testing, where exceptional results are achieved. We then establish a comparative analysis of the performance of our proposed methods with that of SVT to show their superiority and effectiveness.

## 1. Introduction

Brain tumor refers to an abnormal and uncontrollable growth of brain cells, which deprives healthy cells of essential nutrients. It stands as a prominent cause of brain dysfunction and ranks among the most lethal diseases. Broadly classified, brain tumors can be benign or malignant. Benign tumors are characterized by the growth of abnormal cells within the skull, without causing harm to the surrounding brain tissue. Examples of such tumors include meningioma and pituitary tumors. On the other hand, glioma tumors represent a malignant form that poses a grave threat to health and requires immediate attention as a severe case [1].

In order to effectively prevent and treat brain cancer, it is crucial to have a comprehensive understanding of brain tumors and their stages. Diagnostic techniques such as magnetic resonance imaging (MRI) and computed tomography (CT) scans play a vital role in examining and evaluating brain malignancies [2]. Among these, MRIs are the most commonly utilized imaging modality in clinical diagnosis and treatment decision-making for brain tumors [3]. Early detection of brain tumors holds significant importance in medical imaging, as it enables the selection of accurate, precise, and improved treatment methods, ultimately enhancing survival rates and extending life expectancy. By employing advanced imaging technologies, medical professionals can identify brain tumors at their earliest stages, facilitating timely intervention and optimizing patient outcomes.

With the rapid advancement of technology, researchers have begun to explore the potential benefits of combining medical imaging with machine learning techniques to support and enhance the diagnosis and classification of brain tumors. This integration has led to remarkable outcomes, as machine learning-based approaches such as support vector machines (SVM), artificial neural networks (ANN), and pre-trained convolutional neural networks (CNN) like VGG, Xception, Inceptionv3, EfficientNet, Mask RCNN, and CNN have been successfully employed for tumor detection and classification [4].

A notable contribution in this field is the proposal of an optimized convolutional neural network by Irmak et al. [5] specifically designed for tumor detection and classification. Furthermore, Swati et al. [6] developed a transfer learning-based method for tumor detection and classification. It is worth noting that the performance of these models is directly influenced by the size and quality of the dataset available for training [7]. In recent years, majority voting has emerged as a valuable tool to enhance the credibility of classifiers. By considering decisions from multiple classifiers, the approach improves accuracy and reliability compared to relying on a single classifier [8].

Current systems utilized for tumor detection and classification often exhibit limitations in terms of reliability and testing accuracy. To overcome these drawbacks, this research study introduces a novel approach based on bi-fold convolutional neural networks (CNNs). In this study, a total of eighteen machine learning models were trained to detect and classify tumors into their respective categories. Three distinct types of models, namely CNN, VGG-16, and InceptionResNetV2, were employed. For tumor detection, three models were trained for each class, resulting in a total of nine models. Subsequently, the average predictions from each class of models were computed to enhance the overall performance. To further boost the models' performance, accuracy, and robustness, an ensemble-based unsupervised weighted method was employed. By leveraging this innovative technique, the research aimed to address the limitations of existing systems and elevate the performance, accuracy, and reliability of tumor detection and classification. The utilization of bi-fold convolutional neural networks, combined with the ensemble-based weighted approach, holds promise in significantly improving the outcomes of these models.

To cross validate and assess the performance of the model, an averaging technique was employed. Additionally, a bi-fold approach was implemented to integrate the tumor detection and classification models into a single end-to-end connected system. In this methodology, the first model was dedicated to tumor detection. It effectively identified regions of interest or potential tumors within the medical images. Subsequently, only the tumor-identified images were forwarded to the second model. This second model performed the classification task, categorizing the tumors into their respective types, such as meningioma, glioma, and pituitary tumor. By adopting this bi-fold method, the research aimed to establish a comprehensive system that seamlessly combines tumor detection and classification functionalities. This approach enables efficient and accurate identification of tumors and further improves the specificity and precision in classifying them into specific tumor types.

The remaining sections of the paper are organized as follows: Section 2 provides an extensive literature review, focusing on tumor detection and classification techniques and approaches. It presents a systematic overview of relevant research studies and their contributions in these domains. In Section 3, we introduce our proposed novel bi-fold weighted ensemble tumor detection and classification algorithm, outlining its structural design and key components.

Section 4 presents the comparative analysis, results, and discussions, highlighting the major findings and notable performance features of our proposed scheme. Finally, we conclude this research investigation in Section 5 with concluding remarks, summarizing the key insights and contributions of our study.

## 2. Literature review

The advancements in computer-aided systems and biomedical informatics have significantly aided medical professionals through the use of Deep Learning (DL) and Machine Learning (ML) algorithms. Many researchers have developed systems with a claim of high accuracy in tumor detection. However, there is always room for improvement in these studies. In our analysis, we aim to evaluate the performance and accuracy of existing systems.

One approach used for tumor detection involves the utilization of a Fuzzy C-Means (FCM) based method for segmentation, followed by the application of a multilayer Discrete Wavelet Transform combined with Artificial Neural Networks (ANN) for tumor classification. This model achieved an impressive accuracy of 96.97% [9]. Another method employed for brain tumor detection is the implementation of a Convolutional Neural Network (CNN) model combined with VGG-16 architecture. This approach achieved a validation accuracy of 97.16% and a testing accuracy of 91.9% [10]. Furthermore, a deep Convolutional Neural Network Fusion Support Vector Machine algorithm (DCNN-F-SVM) has been utilized for brain tumor detection and segmentation. In this approach, CNN is employed for label prediction, which is then integrated with SVM to form a deep classifier [11].

For the detection of brain tumor, ANN and CNN has been implemented and compared with each other where CNN has achieved better accuracy of 89% [12]. ResNet combined with NASNet, VGG, Xecption, and DenseNet has been applied where the model attains an accuracy of 96% for binary classification of tumor and no-tumor [13]. A CNN plus ResNet, AlexNet, and VGG was proposed to detect tumor alongside data augmentation techniques [14]. Eight layers custom built CNN has been implemented to classify brain tumor into tumor and no-tumor on a small dataset of 253 MRIs which achieved a validation accuracy of 100%. The proposed model has also been compared with transfer-learning techniques such as VGG-16, ResNet50, and Inceptionv3 [15].

To identify tumors and no-tumors, the researchers recommended a CNN, VGG-16 & 19 to extract features more precisely; they also used the deconvolution method on the VGG-16 model, followed by CRF-RCNN on the top layer in place of fully convelutional network (FCN). Accuracy levels for the suggested models VGG-16 and VGG-19 were 95% and 96%, respectively (Sai et al., 2021). Three custom-built CNNs were implemented where first detects tumor and no-tumor with an accuracy of 99.33%. Whereas second classifies no-tumor, meningioma, glioma, and pituitary tumor with a precision of 92.66%. While the last model has been trained to classify tumor into I, II, III, and IV grade with an accuracy of 98.14% [5]. Two models have been proposed by using SVM with genetic algorithm (GA), and ANN with GA resulting in GA-SVM and GA-ANN respectively to detect tumor and no-tumor. Where GA-SVM increased the accuracy of SVM from 79.3% to 91% and GA-ANN increased the accuracy from 75.6% to 94.9% [16].

A classification of no-tumor, meningioma, glioma, and pituitary has been done using the transfer learning model of GoogleNet which attained an accuracy of 98% [17]. A segmentation base grading system was proposed by Sajjad et al. [18] to identify the affected region of the brain using cascade CNN, and then augmentation of the segmented images is done to increase the performance of the VGG-19 model that achieved an accuracy of 90.67% . A custom-designed CNN model has been proposed for the multi-classification of the brain tumor into no-tumor, glioma, meningioma, and pituitary, resulting in 99% validation accuracy [19].

A transfer learning mobileNetV2 has been implemented for feature extraction alongside three feedforward algorithms namely; extreme learning machine, Schmidt neural network, and random vector functional-link networks, using publicly available Harvard dataset (Atlas), attaining the accuracies of 89.24% validation, 91.7% training, and 89% testing case respectively [20]. CNN-based tumor detection and segmentation model has been implemented which achieved an accuracy of 91%, which has been programmed using MATLAB [21].

Younis et al., [22] proposed an ensemble-based technique that was applied to detect brain tumor using CNN and the transfer learning model of VGG-16. The individual accuracies of these models are 96%, and 98.5% respectively while by combining them their final accuracy reaches 98.14%. For the classification of brain tumor a hybrid approach was introduced by Kaur et al. [23], where an independent component analysis (ICA) is applied for extracting the features from MRIs. Furthermore, to optimize the solution hybrid approaches of firefly, cuckoo search, loin, and bat optimization is used and then recurrent neural network (RNN) has been implemented for the classification which achieved an accuracy of 98.61%.

In traditional machine learning, a single algorithm learns a hypothesis from training data. However, ensemble learning takes a different approach by utilizing multiple learners, often referred to as base learners. These base learners are trained on similar types of problems and their predictions are combined to make more accurate predictions [24]. An individual classifier, trained with a specific architecture, may not achieve the same level of accuracy as a group of classifiers working together. Therefore, researchers have explored various approaches to leveraging ensemble models for classification tasks. When different classification models are applied to the same dataset, they can achieve varying levels of accuracy. By combining the predictions of these models, the overall performance and accuracy can be improved compared to the individual models [25].

The utilization of ensemble models in machine learning allows for leveraging the diversity and strengths of multiple classifiers, leading to enhanced predictive capabilities and improved accuracy. Researchers continue to explore and develop innovative approaches to effectively harness the power of ensemble learning in various domains.

In the following table 1, we demonstrate a systematic representation of the existing research work and contributions to highlight the research gap and the strength of the proposed method.

Table 1. A logical comparison table and rational for the research gap

| **Work** | **Method** | **Model** | **Training Approach** | **Testing Approach** | **Expertise** | **Results** |
|---|---|---|---|---|---|---|
| [17] | SVM and KNN-based algorithm | GoogleNet | Best parms training | Taken from training data 70%, 50%, 25% | Classification | 98% |
| [20] | Chaotic Bat algorithm | MobileNetV2, ELM, SNN, and RVFL | Best parms training | Five-fold cross-validation | detection | 96% |
| [26] | block-wise fine-tune CNN | VGG-19 | Block-wise | Five-fold cross-validation | classification | 94.82% |
| [27] | Pipeline-based | ResNet50 | Best parms | Nil | classification | 99% |
| | residual network | | training | | | |
| [28] | Self-defined model | CNN + ANN | Best parms training | Nil | detection | 89% |
| [5] | Three self-defined CNN | CNN | Best parms training | ROC | Detection and classification | 99.33% and 92.66% |
| [18] | Segmentation-based fine-tuned algorithm | VGG-19 | Best parms training | Testing | Grading system | 96.12% |
| [15] | Marker-base watershed, chi-square max | SVM | Best parms training | 10-fold cross-validation | Detection and classification | 98.88% |
| [19] | Custom-built CNN | CNN | Best parms training | Nil | classification | 99% |
| **Proposed method** | Novel bi-fold weighted ensemble method | CNN | Best parms training | Blind testing | Detection & classification | 99.78% & 98.12% |

## 3. Proposed Methodology

This section has been dedicated to demonstrate the proposed methodology for brain tumor detection and classification. We also introduce the datasets, the experimental tools, along with an unsupervised novel weight calculation method using an ensemble learning technique.

### 3.1 Dataset

The dataset used in this research is a combination of three datasets: figshare, SARTAJ, and Br35H. It includes a brain tumor dataset obtained from figshare, which consists of three folders (meningioma, glioma, and pituitary) with a total of 3064 weighted images. Additionally, the second dataset contains 926 gliomas, 937 meningiomas, 896 pituitaries, and 500 no-tumor images. The third dataset contributes 1500 images each to two folders: tumor and no tumor [29 - 31].

### 3.2 Experimental Tools

The experimental setup for our proposed schema to detect and classify brain tumor has been explained in this sub-section. The transfer learning-based models and custom-built CNN have been constructed using tensor-flow and Keras with versions 2.9.1 and 2.9 along with Python version 3.9.6. This experiment has been carried out on Google Colab GPU with the NVIDIA Tesla K80, Cores: 2,496, RAM: 12 GB GDDR5, memory bandwidth: 240 GB/s, Compute Capability: 3.7.

### 3.3 System diagram of the proposed method

The proposed system has been developed by connecting two ends-to-end deep learning averaged weighted ensemble models (AWEMs). In which the first ensemble model has been used to detect the tumor. Whereas, the second ensemble model has been used for the multi-classification of that identified tumor MRI into its exact type/category namely, glioma, meningioma, and pituitary respectively. The below figure 1 shows the basic structure of the proposed system and its working mechanism.

The proposed scheme is a novel Bi-fold (which can be considered as a tree of thoughts as nowadays we are more focus on the logical decision of the deep learning algorithms) tumor detection and classification framework that contains three major layers. The first is the pre-processing layer that loads input MRI data and weighted files of the trained models and then converts the image data into a form that can be used to get predictions from the models. The second layer of the proposed approach focuses on tumor detection from MRIs and aims to differentiate between tumor and no-tumor regions. This is achieved by employing an AWEM, which combines the predictions of multiple models to make more accurate tumor detection decisions.

There are three different techniques used to construct the AWEMs, namely the User-defined weighted method (UWM) and our proposed extended soft voting method (ESVM) and Novel weighted method (NWM). These AWEMs are explained later in the section by using basic numerical illustrations. The tumored MRIs identified in the previous step are then passed through these AWEMs for classification. These classification models further analyze the tumor images and classify them into their specific categories or subtypes. Moreover, the last layer has been structured to demonstrate the results and diagnosis of the exact type of tumor.

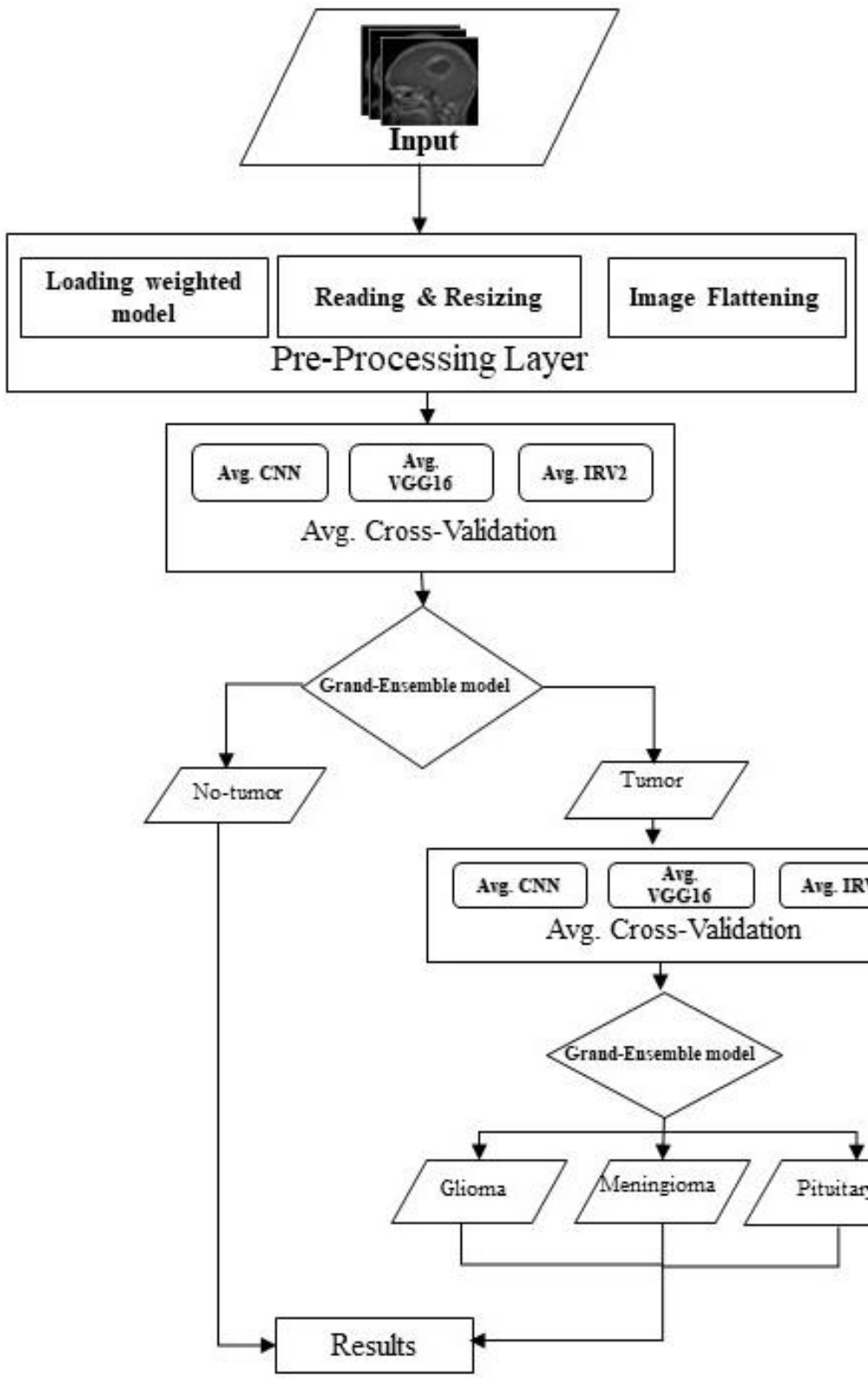

Figure 1 System diagram of the proposed scheme

Figure 2 demonstrates the architecture of the proposed models; where in the left block is the tumor detection model and the block on the right-hand side is the tumor classification model.

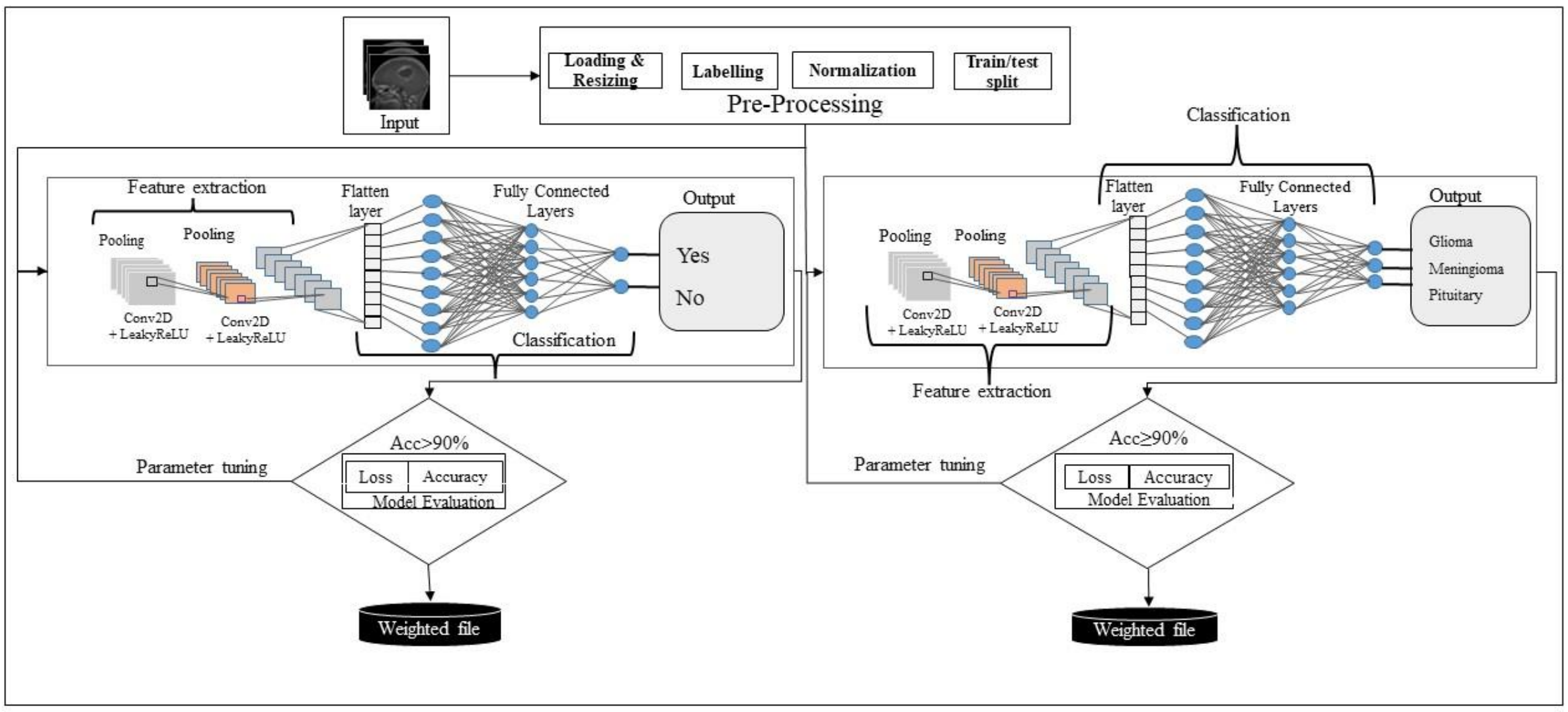

Figure 2. Architecture of the detection and classification models

Three models have been used in this study, which are custom-built CNN, VGG-16, and Inception ResNetV2. The training process starts from the pre-processing layer where the images are converted into a form that can be understandable for the neural network models. The MRIs have been resized from 512×512 to 80×80 because it helps to reduce computational cost and memory. The dataset has been then labelled to represent individual classes, and after that train test split divides the dataset into 80:20 ratios. Lastly, the min-max scalar has been used to convert data into the normalized form.

In the next step, the models have been trained and evaluated on the basis of validation accuracy where the condition of more than 90% has been set as a threshold by avoiding the over-fitting and under-fitting of the model. If the condition is satisfied, the model weighted file would be saved otherwise, readjust the parameter tuning and re-train the model. The same process would be followed for tumor classification models as well. These models were trained for 20 epochs. Furthermore, every model is trained three times which resulted in three different models of the same category.

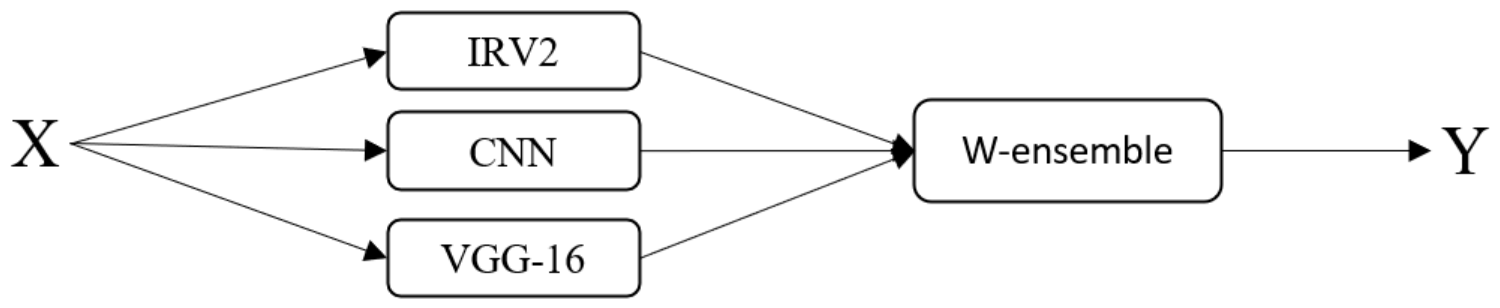

Figure 3 Ensemble learning

The predictions which are made by combining the performance of multiple learners or models are known as voting ensembles. Ensemble learning is committee-based learning and classification technique that is used in ML and DL. Figure 3 shows the basic architecture of the ensemble learning used in this study. There are two types of ensemble learning, namely homogeneous and heterogeneous. The ensemble technique that combines the same type of models is known as homogeneous e.g., regression-based classification. Whereas, the heterogeneous ensemble technique integrates multiple types of models such as CNN and SVM combination-based systems [24]. These systems help in making much stronger decisions as compared to single classifiers. An ensemble learner has a significant advantage in enhancing the ability of weak learners and converting them into strong learners by increasing their classification accuracy [32].

The major research contribution of this study is to propose a new automatic weights calculation method which has been implemented further in the ensemble method. Generally, there are two types of voting approaches in the classification problems which are soft and hard voting. In soft voting, we combine all the probabilities of prediction in each model and then select the prediction with the highest value. Whereas, in the hard voting method the prediction with the highest number of votes is considered as the final result.

Every model/learner has its unique features and limitations, giving it a unique position and relative importance as compared to the others. So, there is a rational need of assigning every model/learner an optimal weight as per its relative importance to enhance the accuracy of the combined effect/result. So, this method has been designed by integrating the concept of soft voting and the principles of Multi-criteria decision-making. Thus, we use our proposed unsupervised weight calculation method to assign optimal weights to every individual model as per its relative importance.

In order to highlight the importance/impact of weights associated with model/individual learners on the final decision, we consider the following intuitive example. In which three models have been considered to predict a particular instance x and their respective prediction accuracies are presented in the following.

$$\text{Model}_1(x) \rightarrow [0.8, 0.2]$$

$$\text{Model}_2(x) \rightarrow [0.7, 0.3]$$

$$\text{Model}_3(x) \rightarrow [0.1, 0.9]$$

Initially, we consider all of the three models are of equal importance and assigned those uniform weights. So, the probability for the x to be belonging to the class zero and class one would be respectively as:

$$\mathrm{p}(j_0 \mid \mathrm{x}) = (0.8 + 0.7 + 0.1)/3 = 0.5333$$

$$\mathrm{p}(j_1 \mid \mathrm{x}) = (0.2 + 0.3 + 0.9)/3 = 0.4666$$

By applying the soft voting principle defined in equation (1), we can identify the category of the instance x having higher probability, which has been explained in the following.

$$\check{p} = arg\ max\ [\mathrm{p}(j_0 \mid \mathrm{x}), \mathrm{p}(j_1 \mid \mathrm{x})] \qquad (1)$$

$$\check{p} = arg\ max\ [\mathrm{p}(j_0 \mid \mathrm{x}), \mathrm{p}(j_1 \mid \mathrm{x})]$$

$$\check{p} = arg\ max\ [0.5333{,}0.4666]$$

$$\check{p} = 0$$

Which indicates that the instance x belongs to the zero category.

Now we consider the individual learners/model as of different relative importance and thus assign their weights as [0.2, 0.2, 0.6] and calculate the probabilities for the instance x as:

$$p(i_0|x) = (0.2\times0.8) + (0.2\times0.7) + (0.6\times0.1) = 0.36$$

$$p(i_1|x) = (0.2\times0.2) + (0.2\times0.3) + (0.6\times0.9) = 0.64$$

Using the equation (1), we choose the class of instance x that has a higher probability.

$$\check{p} = arg\ max\ [p(j_0 \mid x), p(j_1 \mid x)]$$

$$\check{p} = arg\ max\ [0.36{,}0.64]$$

$$\check{p} = 1$$

This approach is termed as user defined weighted method (UWM), which is based on the principle of the soft voting technique. The above prediction result determined by UWM shows that the actual class of the instance x is class one, which is completely against the previous prediction (which has been identified by an equal-weighted method). Thus, from this illustration, we can conclude that the optimal weights play a key role in final decision-making/predictions.

However, the question is how to calculate the optimal weights for the individual model. Because, the accuracy of the user-defined weights is completely subject to the capability, knowledge base and experience of the decision-maker. So, there should be an automated mechanism that could quantify and assign an optimal weight to the individual learner as per its relative importance. Thus, in order to solve this problem we propose an unsupervised method that automatically calculates the weights for each individual learner/model based on its prediction performance using $F_1$-score for every class on the training validation results in the following.

**3.4 Unsupervised Weight calculating Schema.**

Suppose we have three models represented in the form a set as $M = \{M_1, M_2, M_3\}$ and their average $F_1$-scores for the detection and classification of the brain tumor respectively are denoted by a sets as $F_1^d = \{F_1^d(M_1), F_1^d(M_2), F_1^d(M_3)\}$,and $F_1^c = \{F_1^c(M_1), F_1^c(M_2), F_1^c(M_3)\}$. The mathematical form of our proposed unsupervised weight calculation scheme for the models about detection and classification are respectively furnished as follows:

$$W_d(i) = \frac{F_1^d(M_i)}{\sum_{i=1}^{3} F_1^d(M_i)} \quad (2)$$

and $$W_c(i) = \frac{F_1^c(M_i)}{\sum_{i=1}^{3} F_1^c(M_i)} \quad (3)$$

**3.5 Proposed extended weighted soft voting technique**.

In this proposed notion, we improve the idea of soft voting technique by utilizing our newly defined UWCSs defined in equations (2) and (3) to quantify the weights of the models under consideration as per their relative importance based on $F_1$-scores. Then these weights are multiplied with their respective model predictions to calculate the total probabilities of the instances to be belonging to respective classes. We then integrate these resultant values from all of the models to quantify the final prediction using the principle of soft voting technique defined in equation (1). This is working mechanism of our proposed ESVT. The mathematical representation for various steps of this schema can be elaborated as follows. Initially we calculate the probabilities for class zero and class one respectively as

$$p_d(j_0 \mid \mathrm{x}) = \sum_{i=1}^{3} P_i(d) \times W_d(i) \quad (4)$$

$$p_d(j_1 \mid \mathrm{x}) = \sum_{i=1}^{3} P_i(d) \times W_d(i) \quad (5)$$

Thus, the final prediction using ESVT, can be obtained by

$$\breve{p_d} = arg\ max\ [p_d(j_0 \mid \mathrm{x}), p_d(j_1 \mid \mathrm{x})] \quad (6)$$

**3.6 Proposed novel weighted method**.

In this subsection, we demonstrate the proposed novel weighted method, abbreviated as NWM, in which we utilize the product of prediction values of the three models and their respective weights mentioned in equation (4) and (5). The final predictions obtained using our proposed NWM can be expressed mathematically as

$$\breve{p_d} = max[P_i(d) \times W_d(i)] \tag{7}$$

$$\breve{p_c} = max[P_i(c) \times W_c(i)] \tag{8}$$

The pseudo-code for the proposed algorithm has been presented in the following.

**Step 1:**Initialization

**Step 2:** Taking inputs$[P_i(d), P_i(c)\ , F_1^d(M_i), F_1^c(M_i)]$

Where$P_i(d), P_i(c)$ indicates respectively as the predictions for detection and classification with $i = 1,2,3$.

**Step 3:** Calculation of average $F_1$-scores

$$\sum_{i=1}^{3} F_1^d(M_i)\ /3 \text{ , and } \sum_{i=1}^{3} F_1^c(M_i)\ /3$$

**Step 4:** Quantifying weights for detection & classification

$$W_d(i) = \frac{F_1^d(M_i)}{\sum_{i=1}^{3} F_1^d(M_i)} \text{ , and}$$

$$W_c(i) = \frac{F_1^c(M_i)}{\sum_{i=1}^{3} F_1^c(M_i)} \text{ , where } (i = 1, 2, 3)$$

**Step 5:** Integration of weighted predictions

**(a) Extended weighted soft voting method**

Calculation of probabilities for detection

$$p_d(j_0|x) = \sum_{i=1}^{3} P_i(d) \times W_d(i)$$

$$p_d(j_1|x) = \sum_{i=1}^{3} P_i(d) \times W_d(i)$$

Final prediction for the detection

$$\breve{p_d} = arg\ max\ [p_d(j_0 \mid x), p_d(j_1 \mid x)\ ]$$

Calculation of probabilities for the classification

$$p_c(j_0|x) = \sum_{i=1}^{3} P_i(c) \times W_c(i)$$

$$p_c(j_1|x) = \sum_{i=1}^{3} P_i(c) \times W_c(i) \qquad \text{where } (i = 1, 2, 3)$$

$$p_c(j_2|x) = \sum_{i=1}^{3} P_i(c) \times W_c(i)$$

Final prediction for the classification

$$\breve{p_c} = arg\ max\ [p_c(j_0 \mid x), p_c(j_1 \mid x), p_c(j_2 \mid x)]$$

**(b) Novel weighted method**

$$\breve{p_d} = max[P_i(d) \times W_d(i)]$$

$$\breve{p_c} = max[P_i(c) \times W_c(i)]$$

**Step 6: (a)**

```
for j in n:
    if (p̆_d[j][0] > p̆_d[j][1]):
        if (p̆_c[j][0] > p̆_c[j][1] > p̆_c[j][02]):
            predict (glioma)
        elif (p̆_c[j][1] > p̆_c[j][0] > p̆_c[j][02]):
            predict (meningioma)
        else:
            predict(pituitary)
    else:
        predict (notumor)
```

**(b)**

```
for j in n:
    If (p̆_d == 0 || 2 || 4):
        If (p̆_c == 0 || 3 || 6):
            Predict(glioma)
        elif(p̆_c == 1 || 4 || 7):
            Predict (meningioma)
```

else:

Predict (Pituitary)

else:

Predict (notumor)

**Step 7:** End

## 4. Results & Discussion

The multi-criteria decision-making techniques along with the deep learning models have been used in this study to improve the detection and classification accuracy of the ensemble system. Three models have been used in this study namely, custom-built CNN, VGG-16, and InceptionResNetV2. For the sake of cross-validity, every model has been trained three times (where Model1 means when model one has been trained for the first time and its weighted file has been saved, Model2 means trained second time, and Model3 trained for the third time), resulting in nine models in total. The corresponding validation accuracy results of these models on the brain tumor detection dataset have been presented in the following table.

Table 2 Respective accuracies of the models for tumor detection

| Technique | Model1 | Model2 | Model3 |
|---|---|---|---|
| **CNN** | 99.13% | 98.72% | 98.89% |
| **InceptionResNetV2** | 98.84% | 98.60% | 98.72% |
| **VGG-16** | 99.42% | 98.95% | 99.07% |

Moreover, to classify the tumors into its respective type, these models have also been trained and their corresponding validation accuracy results on the brain tumor dataset for classification is presented in the following Table 3.

Table 3 Validation accuracies on training dataset for tumor classification

| Technique | Model1 | Model2 | Model3 |
|---|---|---|---|
| **CNN** | 98.10% | 96.59% | 95.98% |
| **InceptionResNetV2** | 94.47% | 94.77% | 94.54% |
| **VGG-16** | 96.36% | 96.97% | 95.45% |

In order to show the superiority of our proposed extended soft voting ensemble method and novel weighted ensemble method over the user defined weighted ensemble method, we present a comparative analysis by showing their performance in the following table 3. In which we assign a user defined weights for the three models as [0.2, 0.2, 0.6], showing the weights of CNN, InceptionResNetV2 and VGG-16, respectively. After implementing the aforementioned three of the approaches, a blind testing-based approach is used for the assessment of these ensemble systems and their comparative performance results are shown in the following table 4.

Table 4 Confusion matrix for tumor detection

| Method | | Tumor | No-tumor |
|---|---|---|---|
| UWM | **Tumor** | 898 | 8 |
| | **No-tumor** | 0 | 510 |
| ESVM | **Tumor** | 901 | 5 |
| | **No-tumor** | 0 | 510 |
| NWM | **Tumor** | 903 | 3 |
| | **No-tumor** | 0 | 510 |

Table 4 demonstrates the blind testing results of our developed systems (ESVT , NWM) and the UWM. Using the obtained results highlighted in the above confusion matrix, we determine the accuracies of the three methods as 99.43 % for the UWM, 99.64% for the ESVM and 99.78% for the NWM. It has been evident from these results that our both of the proposed methods have significantly higher accuracies as compared to UWM. Furthermore, it can also be clearly seen that our proposed NWM has the highest performance among the three schemes under study.

Table 5 Confusion matrix for brain tumor classification using Bi-fold methods

| | | $G_t$ | $M_t$ | $P_t$ | $N_t$ |
|---|---|---|---|---|---|
| UWM | $G_t$ | 285 | 13 | 2 | 0 |
| | $M_t$ | 18 | 279 | 8 | 8 |
| | $P_t$ | 1 | 1 | 298 | 0 |
| ESVM | $G_t$ | 293 | 5 | 2 | 0 |
| | $M_t$ | 9 | 291 | 5 | 1 |
| | $P_t$ | 0 | 0 | 300 | 0 |
| NWM | $G_t$ | 290 | 8 | 2 | 0 |
| | $M_t$ | 11 | 289 | 1 | 5 |
| | $P_t$ | 0 | 0 | 300 | 0 |

Although, these models have very close validation accuracies, yet the results indicate a significant difference in performance. Where $G_t$ has been used for glioma tumor, $M_t$ represents the meningioma tumor, $P_t$ is the short form of pituitary tumor, while $N_t$ shows notumor. After assessing the performance of the three methods under consideration on tumor detection, the bi-fold system has been developed and evaluated for tumor classification utilizing the UWM, ESVM and NWM. We further calculate the accuracies of the aforementioned methods using the results presented in the above confusion matrix. Where in; UWM achieved an accuracy of 94.15 % and that of proposed ESVM is 97.57%, while the proposed MWM has a highest percentage of accuracy rate as 98.12% for both the detection and classification. The results clearly demonstrate that both of our proposed methods exhibit significantly higher accuracies compared to UWM. Moreover, it is evident that our NWM outperforms the other two schemes studied, showcasing the highest performance.

## 5. Conclusion

In conclusion, accurately detecting and classifying deadly diseases is crucial but challenging with a single model or classifier due to potential errors and unpredictability. To address this challenge, researchers have explored different techniques, including majority voting and ensemble-based approaches. However, decision-making within diverse panels poses its own challenges, such as varying skills and weight assignments.

In this study, we have introduced two novel bi-fold weighted ensemble algorithms based on convolutional neural networks (CNNs) to enhance the performance of brain tumor detection and classification. The first proposed method, called the extended soft voting technique (ESVT), improves the soft voting technique by incorporating a novel unsupervised weight calculating schema (UWCS) for better weight assignment. The second method, known as the novel weighted method (NWM), utilizes the proposed UWCS and weighted optimal prediction to further enhance performance.

Both approaches utilize three distinct models, namely a custom-built CNN, VGG-16, and InceptionResNetV2, trained on publicly available dataset comprising no-tumor and tumor classes namely, Glioma, Meningioma, and Pituitary. These models have been trained using 75:20:5 ratio. Where 75% of the dataset was used for training, 20% for the validation and 5% for the testing. Each model is trained multiple times, and their predictions are averaged to generate a single model.

Through blind testing, our proposed ESVT achieved 99.64% accuracy for tumor detection and 97.57% for classification while our second NWM attained 99.78% accuracy for detection and 98.12% for classification. Where Comparative analysis against the traditional soft voting technique (SVT) further confirms the superiority and effectiveness of our bi-fold novel weighted methods in detecting (99.43%) and classifying (94.15%) brain tumors.

The successful blind testing results and the robust performance of our developed algorithms highlight its efficacy as a valuable tool to support medical professionals in the early diagnosis of diseases. These advancements contributed to the improvement of patient care and treatment outcomes in the field of medical diagnostics.